\documentclass[final]{cvpr}

\usepackage{times}
\usepackage{epsfig}
\usepackage{graphicx}
\usepackage{amsmath}
\usepackage{amssymb}
\usepackage{amsthm}
\usepackage{mathtools}

\usepackage[utf8x]{inputenc}
\usepackage{color}
\usepackage{xcolor}
\usepackage{booktabs,siunitx} 
\usepackage{makecell}
\usepackage{enumitem}
\usepackage[percent]{overpic}
\usepackage[caption=false]{subfig}
\usepackage[noadjust,nocompress]{cite}
\usepackage{rotating}
\usepackage[misc]{ifsym}

\usepackage{multirow}

\usepackage{tabularx}
\usepackage{algorithm}
\usepackage[noend]{algpseudocode}
\usepackage{capt-of,etoolbox}
\usepackage[accsupp]{axessibility} 
\usepackage[pagebackref=true,breaklinks=true,colorlinks,bookmarks=false]{hyperref}
\usepackage{mwe}

\if@mathematic
   
\else
   
\fi

\definecolor{road}                {RGB}{128, 64,128}
\definecolor{sidewalk}            {RGB}{244, 35,232}
\definecolor{building}            {RGB}{ 70, 70, 70}
\definecolor{wall}                {RGB}{102,102,156}
\definecolor{fence}               {RGB}{190,153,153}
\definecolor{pole}                {RGB}{153,153,153}
\definecolor{traffic light}       {RGB}{250,170, 30}
\definecolor{traffic sign}        {RGB}{220,220,  0}
\definecolor{vegetation}          {RGB}{107,142, 35}
\definecolor{terrain}             {RGB}{152,251,152}
\definecolor{sky}                 {RGB}{ 70,130,180}
\definecolor{person}              {RGB}{220, 20, 60}
\definecolor{rider}               {RGB}{255,  0,  0}
\definecolor{car}                 {RGB}{  0,  0,142}
\definecolor{truck}               {RGB}{  0,  0, 70}
\definecolor{bus}                 {RGB}{  0, 60,100}
\definecolor{train}               {RGB}{  0, 80,100}
\definecolor{motorcycle}          {RGB}{  0,  0,230}
\definecolor{bicycle}             {RGB}{119, 11, 32}
\definecolor{void}                {RGB}{  0,  0,  0}

\newcommand{\vecnorm}[1]{\left\|#1\right\|}

\newtheorem{definition}{Definition}

\newcolumntype{P}[1]{>{\centering\arraybackslash}p{#1}}

\newcommand{\PAR}[1]{\vskip2pt \noindent{\bf #1}}

\begin{document}

\title{P3Depth: Monocular Depth Estimation with a Piecewise Planarity Prior}

\author{
Vaishakh Patil\textsuperscript{1}\space\space\space\space Christos Sakaridis\textsuperscript{1}\space\space\space\space Alexander Liniger\textsuperscript{1}\space\space\space\space Luc Van Gool\textsuperscript{1,2}\vspace{6px} \\
\textsuperscript{1}Computer Vision Lab, ETH Z\"urich\space\space\space\space \textsuperscript{2}PSI, KU Leuven\\
 }

 \maketitle

\begin{abstract}
\vspace{-0.2cm}
Monocular depth estimation is vital for scene understanding and downstream tasks. We focus on the supervised setup, in which ground-truth depth is available only at training time. Based on knowledge about the high regularity of real 3D scenes, we propose a method that learns to selectively leverage information from coplanar pixels to improve the predicted depth. In particular, we introduce a piecewise planarity prior which states that for each pixel, there is a seed pixel which shares the same planar 3D surface with the former. Motivated by this prior, we design a network with two heads. The first head outputs pixel-level plane coefficients, while the second one outputs a dense offset vector field that identifies the positions of seed pixels. The plane coefficients of seed pixels are then used to predict depth at each position. The resulting prediction is adaptively fused with the initial prediction from the first head via a learned confidence to account for potential deviations from precise local planarity. The entire architecture is trained end-to-end thanks to the differentiability of the proposed modules and it learns to predict regular depth maps, with sharp edges at occlusion boundaries. An extensive evaluation of our method shows that we set the new state of the art in supervised monocular depth estimation, surpassing prior methods on NYU Depth-v2 and on the Garg split of KITTI. Our method delivers depth maps that yield plausible 3D reconstructions of the input scenes. 
Code is available at: \url{https://github.com/SysCV/P3Depth}
\end{abstract}
\vspace{-0.3cm}
\section{Introduction}
\label{sec:intro}

Depth estimation is a fundamental problem in computer vision. It consists in predicting the perpendicular coordinate of the 3D point depicted at each pixel.  Applications range from robotics to autonomous cars. There is experimental evidence~\cite{does:vision:matter:for:action} that depth is the most vital vision-level cue for executing actions, together with semantic segmentation. In this work, we focus on monocular depth estimation, which involves the challenge of scale ambiguity, as the same input image can be generated by infinitely many 3D scenes.

The current trend in solving this task involves fully convolutional neural networks that output a dense depth prediction either with standard supervision on depth~\cite{depth:multiscale:network,depth:frcn,deep:ordinal:regression:network,structure:guided:ranking:loss:depth} or with self-supervision by using the predicted depth to reconstruct neighboring views of the scene~\cite{unsupervised:cnn:single:view:depth,monodepth,unsupervised:depth:ego:motion:from:video,monodepth2}. Most supervised approaches use a pixel-level loss which treats predictions at different pixels separately. This regime ignores the high degree of \emph{regularity} of real-world 3D scenes, which generally yield \emph{piecewise smooth} depth maps.

\begin{figure}
    \centering
    \includegraphics[width=0.85\linewidth]{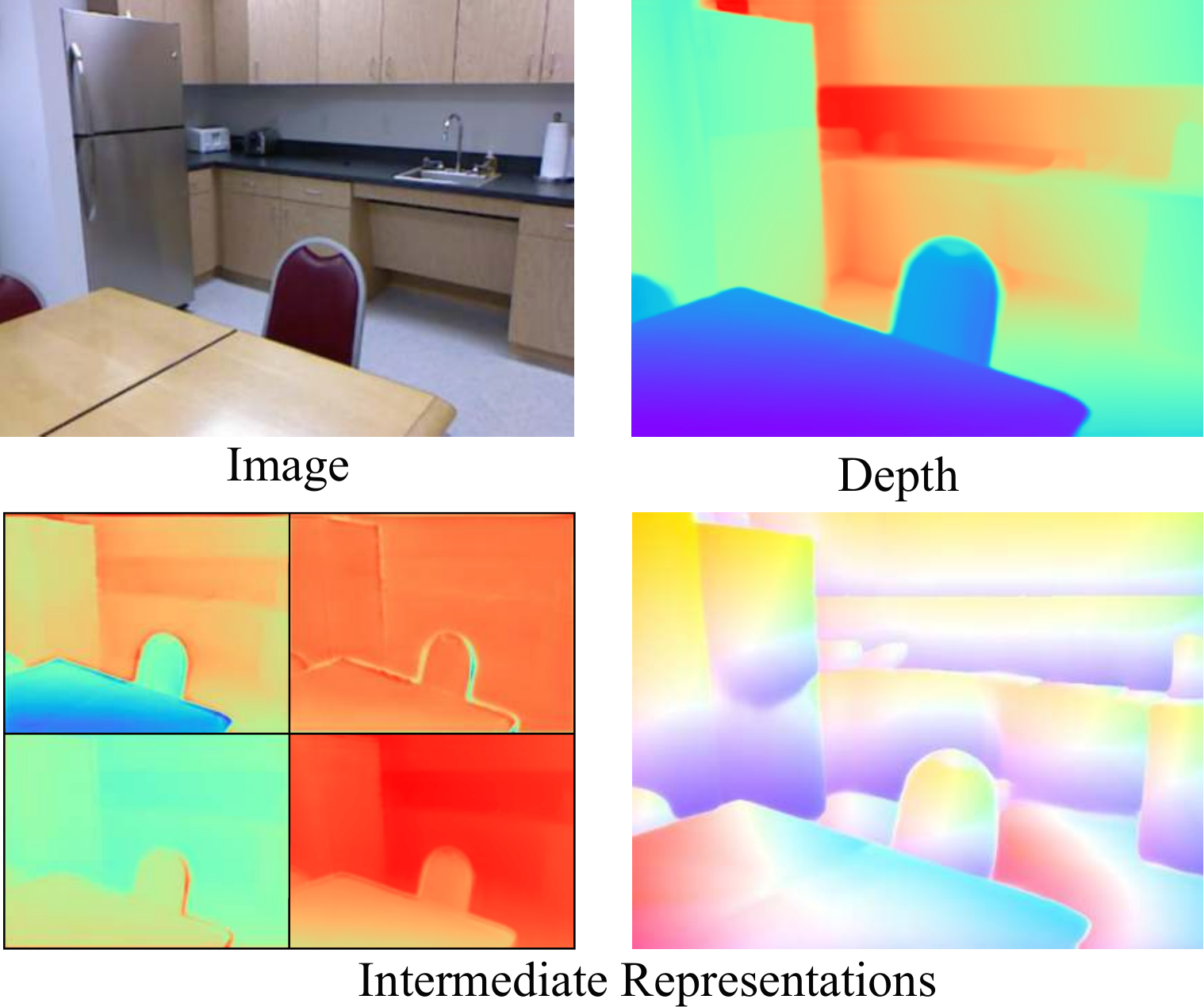}
    \captionof{figure}{Real-world 3D scenes have a high degree of regularity. We propose a method which can exploit this regularity, by implicitly learning intermediate representations that contain useful information about local planes in the scene. The proposed end-to-end model predicts high-quality depth maps with sharp edges at occlusion boundaries, which yield consistent 3D reconstructions.}
    \label{fig:teaser}
    \vspace{-0.4cm}
\end{figure}

A common choice for modeling this prior knowledge of the geometry of real 3D scenes are planes~\cite{robust:piecewise:planar:3d:reconstruction,fast:approximate:piecewise:planar:modeling,planenet:piecewise:planar:reconstruction,planercnn:3d:plane:detection:reconstruction}. Planes are the local first-order Taylor approximation for locally differentiable depth maps and they are easy to parameterize using three independent coefficients. Once a pixel is associated with a plane, its depth can be recovered from the position of the pixel and the coefficients of the associated plane. In~\cite{geolayout}, such a plane coefficient representation is used to learn to predict planes explicitly.

We adopt the plane representation from~\cite{geolayout}, but we depart from the explicit prediction of planes and rather use this representation as an appropriate output space for defining interactions between pixels based on planarity priors. In particular, the first head of our network outputs dense plane coefficient maps which are afterwards converted to depth maps, as shown in Fig.~\ref{fig:main_pipeline}. Predicting plane coefficients is motivated by the fact that two pixels $\mathbf{p}$ and $\mathbf{q}$ that belong to the same plane ideally have equal plane coefficient representations, whereas they generally have different depth. Thus, using the plane coefficient representation of $\mathbf{q}$ for predicting depth at the position of $\mathbf{p}$ results in a correct prediction if the pixels belong to the same plane.

We leverage this property by learning to identify \emph{seed pixels} which share the same plane as the examined pixel, whenever such pixels exist, in order to selectively use the plane coefficients of these pixels for improving the predicted depth. This approach is motivated by a piecewise planarity prior which states that for each pixel $\mathbf{p}$ with an associated 3D plane, there is a seed pixel $\mathbf{q}$ in the neighborhood of $\mathbf{p}$ which is associated with the same 3D plane as $\mathbf{p}$. To predict depth with this scheme, we need to identify (i) the regions where the prior is valid and (ii) the seed pixels in these regions, by predicting the offsets $\mathbf{q} - \mathbf{p}$. We thus design a second head in the network, which outputs a dense offset vector field and a confidence map, as shown in Fig.~\ref{fig:main_pipeline}. The predicted offsets are used to resample the plane coefficients from the first head and generate a second depth prediction. The depth predictions from the two heads are then fused adaptively using the confidence map as fusion weights, in order to down-weigh the offset-based prediction and rely primarily on the basic depth prediction in regions where the piecewise planarity prior is not valid, e.g.\ on parts of the scene with high-frequency structures. Supervision on the offsets and confidence map is applied implicitly, by supervising the fused depth prediction. Thanks to using seed pixels for prediction, our model implicitly learns to group pixels based on their membership in smooth regions of the depth map. This helps preserve sharp depth discontinuities, as shown in Fig.~\ref{fig:teaser}. Last but not least, we propose a mean plane loss which enforces first-order consistency of our predicted 3D surfaces with the ground truth and further improves performance.

We evaluate our method extensively on 6 datasets for supervised monocular depth estimation: NYU Depth-v2, KITTI, ScanNet, SUN-RGBD, DIODE Indoor, and ETH-3D. Comparisons to competing approaches demonstrate that we set a new state of the art on NYU Depth-v2 and KITTI, surpassing the former best-performing method in all commonly used evaluation metrics on NYU and on the Garg split~\cite{unsupervised:cnn:single:view:depth} of KITTI. Moreover, in a challenging zero-shot transfer setup, we outperform the prior state of the art on ScanNet, SUN-RGBD, DIODE Indoor, and ETH-3D. We conduct a thorough ablation study and show quantitatively the merit of our novel formulation for depth prediction. We also provide qualitative comparisons with the prior state of the art, which evidence the high quality of our predictions, in particular when the latter are used for 3D reconstruction.

\section{Related Work}
\label{sec:related}

\PAR{Supervised monocular depth estimation} assumes that ground-truth depth maps are available for training images and requires inference on single images. A notable early approach is Make3D~\cite{depth:make3d}, which explicitly handcrafts a piecewise planar structure on the scene and learns the associated parameters locally using a Markov random field. The multi-scale network of \cite{depth:multiscale:network} pioneered the usage of deep CNNs in depth estimation by learning an end-to-end mapping from images to depth maps. Several works have afterwards focused on this setting, proposing i.a.\ (i) more advanced architectures such as residual networks~\cite{depth:frcn}, convolutional neural fields~\cite{structured:attention:crf,depth:convolutional:neural:fields}, fusion of multiple scales in the frequency domain~\cite{depth:fourier:analysis}, transformer-based blocks that attend to global depth statistics~\cite{adabins} and depth merging networks for handling multiple resolutions~\cite{high:resolution:depth:estimation:multi:resolution:merging}, (ii) losses that are better suited for depth prediction such as reverse Huber loss~\cite{depth:frcn}, classification loss~\cite{depth:as:classification}, ordinal regression loss~\cite{deep:ordinal:regression:network}, pairwise ranking loss~\cite{structure:guided:ranking:loss:depth} and adaptive combinations of several depth-related losses~\cite{multi:loss:rebalancing:monocular:depth}, and (iii) joint learning of depth with normals or semantic labels~\cite{depth:normals:labels,geonet:depth:normal,padnet}. The ambiguity in depth shift and focal length scale in mixed-data setups is addressed in~\cite{3d:scene:shape:from:single:image} by applying 3D point cloud encoders to the lifted depth map. Our method belongs to this category and casts the depth prediction to a more appropriate space for exploiting regularities of input scenes.

\PAR{Other depth estimation setups} include unsupervised and semi-supervised monocular depth estimation as well as stereo-based depth estimation. Unsupervised learning of depth with stereo pairs based on novel view synthesis~\cite{deepstereo} that uses an image reconstruction loss in which the predicted depth is used for warping one image of the pair to the frame of the other was introduced in~\cite{unsupervised:cnn:single:view:depth} and was cast in a fully differentiable formulation in~\cite{monodepth}.
Further works in this direction leverage temporal information~\cite{undemon,unsupervised:monocular:depth:and:odometry,recurrent:depth:estimation:from:monocular:video}. The need for stereo pairs in this framework was lifted in~\cite{unsupervised:depth:ego:motion:from:video}, which operates on monocular videos.
Consistency of the estimated 3D structures and of ego-motion across video frames is enforced in~\cite{unsupervised:depth:ego:motion:3d:geometric:constraints,neural:rgb2d:sensing,robust:consistent:video:depth:estimation}.
Depth and ego-motion are combined with optical flow and semantics in~\cite{geonet:unsupervised,self:supervised:monocular:depth:dynamic:object}
and with edges in~\cite{lego}.
Robustness to occlusions across video frames is achieved in~\cite{monodepth2} with a minimum reprojection loss. The optimization is facilitated with specialized losses in~\cite{feature:metric:loss:depth:egomotion,pladenet:self:supervised:depth}.
Recent methods exploit video input at test time~\cite{temporal:opportunist:depth}, consistency to segmentation outputs~\cite{edge:depth} and scale consistency across adjacent frames~\cite{scale:consistent:monocular:depth}.
Unsupervised approaches generally assume more complex training data than supervised ones and suffer from scale ambiguity and violations of the Lambertian assumption. Semi-supervised depth estimation is introduced in~\cite{semi:supervised:learning:monocular:depth}, which combines sparse depth measurements with an image reconstruction loss. Dataset-specific assumptions on the presence and the format of depth supervision are also relaxed in~\cite{megadepth}, which utilizes multi-view image collections for generating reliable large-scale depth supervision, and in~\cite{midas}, where diverse datasets providing different forms of supervision for monocular depth estimation are leveraged to generalize better on unseen data. 
Early stereo methods rely on hand-crafted matching costs~\cite{semiglobal:matching:stereo} for estimating disparity. Initial approaches that learned the matching function include~\cite{learning:the:matching:function,stereo:matching:convolutional:neural:network}, while subsequent works rely on fully convolutional architectures~\cite{dataset:disparity:optical:flow:scene:flow,psmnet}.
Stereo methods also assume more complex data in the form of stereo pairs both at training and testing, which prevents their application to more general and uncontrolled monocular settings.

\PAR{Geometric priors for depth} have been extensively studied in the literature. In particular, the piecewise planarity prior has been traditionally used in multi-view stereo~\cite{piecewise:planar:and:nonplanar:stereo} and 3D reconstruction~\cite{robust:piecewise:planar:3d:reconstruction,fast:approximate:piecewise:planar:modeling} in order to make these problems amenable to faster optimization. These approaches involve explicit depth planes and fit these planes on image superpixels or point sets from input point clouds. Superpixel-level depth planes are also leveraged in depth denoising and completion~\cite{stereoscopic:inpainting,sfsu:synthetic}. In more recent, deep learning-based approaches, the incorporation of geometric priors is performed either explicitly by segmenting planes~\cite{structdepth,planenet:piecewise:planar:reconstruction,piecewise:planar:3d:reconstruction:associative:embedding,planercnn:3d:plane:detection:reconstruction} or implicitly by properly designing the loss~\cite{p2net:unsupervised:indoor:depth}. Non-local 3D context is leveraged in the virtual normal framework of~\cite{virtual:normal:depth} by using supervision from virtual planes which correspond to triplets of non-collinear points of the depth map. A non-local coplanarity constraint is embedded to the network in~\cite{depth:attention:volume} via a depth-attention volume. 
Surface normals are used in~\cite{adaptive:surface:normal:constraint:depth} to increase geometric consistency on regular structures. A representation directly associated with coefficients of 3D planes in the image space without dependence on intrinsic camera parameters is used in~\cite{geolayout, recovering:planes:eccv:2018} for estimating the dominant depth planes in the scenes. The same representation with plane coefficients is employed in~\cite{local:planar:guidance} to guide the upsampling modules of the decoder part of depth networks, achieving state-of-the-art performance. We also use this representation with plane coefficients, but contrary to~\cite{geolayout,recovering:planes:eccv:2018}, we learn it without requiring annotations for planes. Instead, we optimize plane coefficients together with spatial offset vectors to learn to identify coplanar pixels and use this coplanarity for predicting depth.
While offset vectors are also used in~\cite{occlusion:boundaries:using:displacement:fields} for post-processing depth by merely resampling the prediction, we incorporate offset vectors in a single end-to-end architecture and generate the prediction via interpolation with the plane associated with the seed pixel pointed by the offset. Our approach is loosely inspired by~\cite{instance:segmentation:spatial:embeddings}, which trains offset vectors to identify instance segmentation centers from annotated images, while we focus on depth prediction and operate without supervision for plane instances.

\section{Method}
\label{sec:method}

\begin{figure*}[!t]
    \centering
    \includegraphics[width=\textwidth]{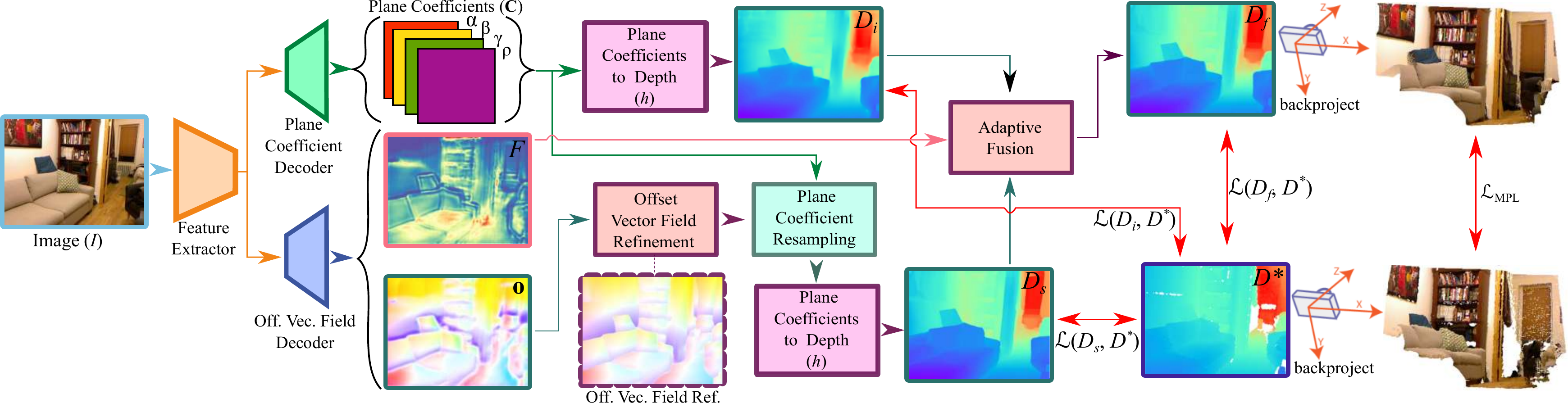}
    \caption{\textbf{Overview of our end-to-end P3Depth method.} P3Depth includes two output heads. The first head outputs pixel-level plane coefficients ($\mathbf{C}$), while the second head outputs a dense offset vector field ($\mathbf{o}$) identifying positions of seed pixels along with a confidence map ($F$). Then, the plane coefficients of seed pixels are used to predict depth at each position. The resulting prediction ($D_s$) is adaptively fused with the initial  prediction ($D_i$) using the confidence map to account for potential deviations from precise local planarity.}
    \label{fig:main_pipeline}
    \vspace{-0.3cm}
\end{figure*}

As pointed out in Sec.~\ref{sec:intro}, our network estimates depth by selectively combining depth from each pixel and its corresponding seed pixel. For this formulation to work, it is vital to use a common representation which can capture pixel-wise depth as well as planarity information. We achieve this by using a plane coefficient representation similar to~\cite{geolayout}. We explain this representation and derive an analytical relation between plane coefficients and depth in Sec.~\ref{sec:method:plane:coefficients}, which allows us to supervise the network only with depth. The main advantage of the plane coefficient representation is that the depth of a pixel in the image can be directly computed by the plane coefficients of a different pixel, under the assumption that the two pixels are on the same plane. This advantage forms the basis of our planarity prior and the selective planar depth bootstrapping using seed pixels, which we explain in Sec.~\ref{sec:method:seed:pixels}. Finally, in Sec.~\ref{sec:method:mean:planes} we present an additional patch-level mean plane loss, which is complementary to the previous components and contributes independently to learning regular depth maps.

\subsection{Preliminaries}
\label{sec:method:preliminaries}

Monocular depth estimation requires learning a dense mapping $f_\theta: I(u, v) \rightarrow D(u, v)$, where $I$ is the input image with spatial dimensions $H \times W$, $D$ is the corresponding depth map of the same resolution, $(u, v)$ are pixel coordinates in the image space and $\theta$ are the parameters of the mapping $f$. In the supervised setup, a ground-truth depth map $D^*$ is available for each image $I$ at training time. During training, the parameters $\theta$ are optimized such that the function $f_\theta$ minimizes the difference between the predicted depth and the ground-truth depth over the training set $\mathcal{T}$. This can be formalized as
\begin{equation}
    \label{eq:learning_depth}
    \min_\theta \; \sum_{(I, D^*) \in \mathcal{T}}\mathcal{L}(f_\theta(I), D^*),
\end{equation}
where $\mathcal L$ is a loss function that penalizes deviations between the prediction and the ground truth. Furthermore, given a depth map $D$ along with the camera intrinsics, we can backproject each pixel to the 3D space. Using the pinhole camera model and given the focal lengths $(f_x, f_y)$ and the principal point $(u_0, v_0)$, every pixel $\mathbf{p} = (u, v)^T$ is mapped to a 3D point $\mathbf{P}=(X,Y,Z)^T$ according to
\begin{equation}
    \label{eq:2D_3D}
    Z=D(u, v),\;X=\frac{Z(u-u_{0})}{f_x},\;Y=\frac{Z(v-v_{0})}{f_y}.
\end{equation}

\subsection{Plane Coefficient Representation for Depth}
\label{sec:method:plane:coefficients}

Suppose that the backprojected 3D point $\mathbf{P}$ corresponds to a planar part of the 3D scene. The equation of the associated plane in the point–normal form can be written as $\mathbf{n} \cdot \mathbf{P}+d=0$, where $\mathbf{n}=(a,b,c)^T$ is the normal vector to the plane and $-d$ is the distance of the plane from the origin. Substituting $\mathbf{P}$ from \eqref{eq:2D_3D} into the point-normal equation yields
\vspace{-0.2cm}
\begin{equation}
    \label{eq:inv_plane}
    \frac{1}{Z} = \underbrace{\frac{-a}{f_{x}d}}_{\hat{\alpha}}u + \underbrace{\frac{-b}{f_{y}d}}_{\hat{\beta}}v + \underbrace{\frac{1}{d}(\frac{a}{f_{x}}u_{0}+\frac{b}{f_{y}}v_{0}-c)}_{\hat{\gamma}}.
\vspace{-0.2cm}
\end{equation}
Thus, for image regions that depict planar 3D surfaces, the inverse depth is an affine function of pixel position, where the coefficients encode both the camera intrinsics and the 3D plane. We reformulate \eqref{eq:inv_plane} by introducing $\rho=\sqrt{{\hat{\alpha}}^2 + {\hat{\beta}}^2 + {\hat{\gamma}}^2}$ and normalizing $\alpha = \frac{\hat{\alpha}}{\rho}$, $\beta = \frac{\hat{\beta}}{\rho}$ and $\gamma = \frac{\hat{\gamma}}{\rho}$ into
\vspace{-0.4cm}
\begin{equation}
    \label{eq:param_depth}
    Z = [(\alpha u + \beta v + \gamma) \rho]^{-1}.
\end{equation}
We term $\mathbf{C} = (\alpha,\beta,\gamma,\rho)^T$ as the \emph{plane coefficients}. Using this notation, \eqref{eq:param_depth} can be written as $Z = h(\mathbf{C}, u, v)$. Instead of directly predicting depth, we design our model to have a plane coefficient head, which first predicts a dense plane coefficient representation $\mathbf{C}(u, v)$ and then applies \eqref{eq:param_depth} to compute an initial depth prediction which we denote with $D_i$. More formally, the mapping $f_\theta$ from Sec.~\ref{sec:method:preliminaries} is now a composition $f_\theta = h \circ (\mathbf{g}_\theta, \mathbf{p})$, where $\mathbf{g}_\theta: I(u, v) \rightarrow \mathbf{C}(u, v)$ maps the input image to the plane coefficient representation and $h: (\mathbf{C}(u, v), u, v) \rightarrow D_i(u, v)$ applies \eqref{eq:param_depth} at each pixel. 
Supervision is applied to the output depth $D_i$ via \eqref{eq:learning_depth}.

Predicting the plane coefficients as an intermediate output does not give a immediate advantage compared to directly predicting the depth. However, two pixels that depict the same 3D plane have the same parameters $\mathbf{C}$, but generally a different depth. This fact is the core to the next part of the network, which allows to predict depth by selectively bootstrapping the plane coefficients from a seed pixel. 

\subsection{Learning to Identify Seed Pixels}
\label{sec:method:seed:pixels}

Let us assume we have one pixel $\mathbf{p}$ which belongs to a planar surface in 3D. By definition, every other pixel on this planar surface has the same $\mathbf{C}$ values. Thus, in an ideal world the network only has to predict $\mathbf{C}$ at one of these pixels, $\mathbf{q}$, to get all their depth values correct. This pixel can be interpreted as the seed pixel that describes the plane. However, defining this seed pixel and the region in which the depth should be bootstrapped from it is non-trivial. Thus, in this work we let the network discover this seed pixel and the respective region.

Formally, let us start by defining our piecewise planarity prior which is a relaxed version of the previous idea.
\begin{definition} \label{def:piecewise:planarity:prior}
\textbf{(Piecewise planarity prior)} For every pixel $\mathbf{p}$ with an associated 3D plane, there exists a seed pixel $\mathbf{q}$ in the neighborhood of $\mathbf{p}$ which is also associated with the same plane as $\mathbf{p}$.
\end{definition}
Note that in general, there may exist multiple seed pixels or no seed pixel for $\mathbf{p}$.

Given that the prior holds, the task of depth prediction for $\mathbf{p}$ can also be solved by identifying $\mathbf{q}$, i.e., by predicting the offset $\mathbf{o}(\mathbf{p}) = \mathbf{q} - \mathbf{p}$. Thus, we design our model so that it features a second, offset head and let this offset head predict a dense offset vector field $\mathbf{o}(u, v)$. The two heads of the network share a common encoder and have separate decoders, as shown in Fig.~\ref{fig:main_pipeline}. We use the predicted offset vector field to resample the plane coefficients via
\begin{equation} \label{eq:resampling}
    \mathbf{C}_s(\mathbf{p}) = \mathbf{C}(\mathbf{p} + \mathbf{o}(\mathbf{p})),
\end{equation}
using bilinear interpolation to handle fractional offsets. The resampled plane coefficients are then used to compute a second depth prediction 
\begin{equation} \label{eq:depth:seed}
    D_s(u, v) = h(\mathbf{C}_s(u, v), u, v),
\end{equation}
based on the seed locations. This allows the network to bootstrap the depth from the seed pixel. 

\begin{figure}[!t]
\includegraphics[width=\linewidth]{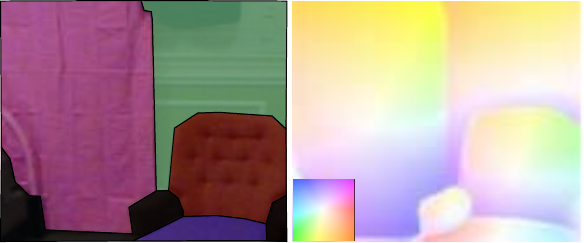}
\caption{\textbf{Ground truth planes vs. predicted offset vector field.} The predicted offset vector at a pixel tends to point towards a seed pixel with which it shares the same plane coefficients. The left image shows the overlayed labels of the segmented planes on an example from NYU Depth-v2 and the right image shows the respective predicted offset vector field. The bottom left legend in the right image shows the color coding for the vector field.}
\label{fig:flow_fig}
\vspace{-0.3cm}
\end{figure}

However, the prior is not always valid, so the initial depth prediction $D_i$ may actually be preferable compared to the seed-based prediction $D_s$. To account for such cases, the second head additionally predicts a confidence map $F(u, v) \in [0,\,1]$, which indicates the confidence of the model in using the predicted seed pixels for estimating depth via $D_s$. The confidence map is leveraged to compute the final depth prediction by adaptively fusing $D_i$ and $D_s$:
\begin{equation} \label{eq:depth:fused}
    D_f(u, v) = F(u, v)D_s(u, v) + (1 - F(u, v))D_i(u, v).
\end{equation}
We apply supervision on each of $D_f$, $D_i$ and $D_s$ in our model, by optimizing the following loss:
\begin{equation}
    \mathcal{L}_{\text{depth}} = \mathcal{L}(D_f, D^*) + \lambda\mathcal{L}(D_s, D^*) + \mu\mathcal{L}(D_i, D^*),
\end{equation}
with $\lambda$ and $\mu$ being hyperparameters. In this way, we encourage (i) the plane coefficient head to output a representation that is accurate across all pixels even when they have a high confidence value and (ii) the offset head to learn high confidence values for pixels for which the planarity prior holds and low confidence values for the converse.

However, there is a caveat in this formulation. In particular, the model is not supervised directly on the offsets. In fact, it could simply predict zero offsets everywhere and still produce valid predictions $D_s$ and $D_f$, which would be identical to $D_i$. This unwanted behavior is avoided in practice thanks to the fact that the initial predictions $D_i$ are erroneously smoothed near depth boundaries, due to the regularity of the mapping $f_\theta$ for the case of neural networks. As a result, for pixels on either side of a boundary, predicting a non-zero offset that points away from the boundary yields a lower value for $\mathcal{L}_{\text{depth}}$, because such an offset uses a seed pixel for $D_s$ which is further from the boundary and suffers from smaller error owing to smoothing. Also due to regularity of the mapping that generates the offset vector field, these non-zero offsets are propagated from the boundaries to the inner parts of regions with smooth depth, helping the network to predict non-trivial offsets.

In the fully-fledged version of our method, we cascade the offset vectors multiple times before resampling the plane coefficient maps. For example, a single cascading step samples the position $\mathbf{p} + \mathbf{o}(\mathbf{p}) + \mathbf{o}(\mathbf{p} + \mathbf{o}(\mathbf{p}))$. Our motivation for this cascaded refinement is that seed pixels within the same planar region should converge to the center of the region, which helps accumulate information from more pixels in predicting the plane coefficients of the region. At the same time, pixels without a reliable seed pixel are anyway assigned a low confidence value, so cascading the offsets does not hurt the respective depth prediction.

\subsection{Mean Plane Loss}
\label{sec:method:mean:planes}

The assumption we use for formulating our mean plane loss is that given a pixel coordinate, its neighboring pixels should lie on the same plane in the 3D space. The normal $\mathbf{n}$ of this plane should satisfy an overdetermined system of linear equations. However, ground-truth depth maps are usually captured by consumer-level sensors with noisy measurements and limited precision, which renders the above regime for local fitting of normals inapplicable, as finding the true optimal solution is not guaranteed.

Even though this is a valid observation, depth measurements still contain comprehensive details about the scene structure. This information can be aggregated locally to enforce first-order consistency between the predicted and the ground-truth 3D surface. Normals are one way how this aggregation across patches can be performed. For an input patch, the corresponding normal $\mathbf{n}$ needs to satisfy $\mathbf{A} \mathbf{n} = \mathbf{b}$, s.t. $\vecnorm{\mathbf{n}}_{2}=1$, where $\mathbf{A}$ is a data matrix build by stacking the 3D points in the patch and $\mathbf{b}$ is a vector of ones. Following~\cite{data-driven:3D:ICCV:2013, geonet:depth:normal}, the closed-form solution of this least-squares problem is:
\vspace{-0.3cm}
\begin{align}
\small
    \label{eq:least_square}
    \mathbf{n} = \frac{{(\mathbf{A}^{T}\mathbf{A})}^{-1}\mathbf{A}^{T}\mathbf{b}}{\vecnorm{{(\mathbf{A}^{T}\mathbf{A})}^{-1}\mathbf{A}^{T}\mathbf{b}}_{2}}.
\vspace{-0.1cm}
\end{align}
To compute the mean plane loss, we first estimate surface normals for all $K$ non-overlapping patches in $D$ and $D^{*}$ and then penalize their difference via
\vspace{-0.3cm}
\begin{align}
    \label{eq:MPL}
    \mathcal{L}_{\text{MPL}} = \sum_{k=1}^{K} \vecnorm{\mathbf{n}_k - \mathbf{n}^{*}_k}_1.
\end{align}

For patches with depth discontinuities, even when the $\mathbf{n}_k^*$ for patch $k$  does not correspond to a ground-truth 3D plane, the mean plane loss still provides a useful supervision signal, as it penalizes local depth profiles that are inconsistent with $\mathbf{n}_k^*$. Also, we do not require ground-truth normals, as opposed to~\cite{geonet:depth:normal}. Given \eqref{eq:least_square}, we can see that the loss \eqref{eq:MPL} directly affects the depth of all points inside the patch via $\mathbf A$. Finally, the complete loss is $\mathcal{L}_{\text{total}} = \mathcal{L}_{\text{depth}} + \mathcal{L}_{\text{MPL}}$.

\begin{figure}[!t]
\begin{minipage}{\columnwidth}
    \noindent\begin{minipage}{0.1\columnwidth}
        \footnotesize
        \begin{tabular}{l}
        \\ [-0.4em]
        \rotatebox{90}{\textbf{Image}}\\
        \\ [-0.4em]
        \rotatebox{90}{\textbf{DORN}~\cite{depth:ordinal:cvpr:2018}}\\
        \\ [-0.6em]
        \rotatebox{90}{\textbf{VNL}~\cite{virtual:normal:depth}}\\
        \\ [-0.8em]
        \rotatebox{90}{\textbf{BTS}~\cite{local:planar:guidance}}\\
        \\ [0.2em]
        \rotatebox{90}{\textbf{Ours}}\\
        \\ [1.0em]
        \rotatebox{90}{\textbf{GT}}\\
        \\  
        \end{tabular}
     \end{minipage}
     \hspace{-0.5cm}
     \noindent\begin{minipage}{0.9\columnwidth}
            \includegraphics[width=75mm]{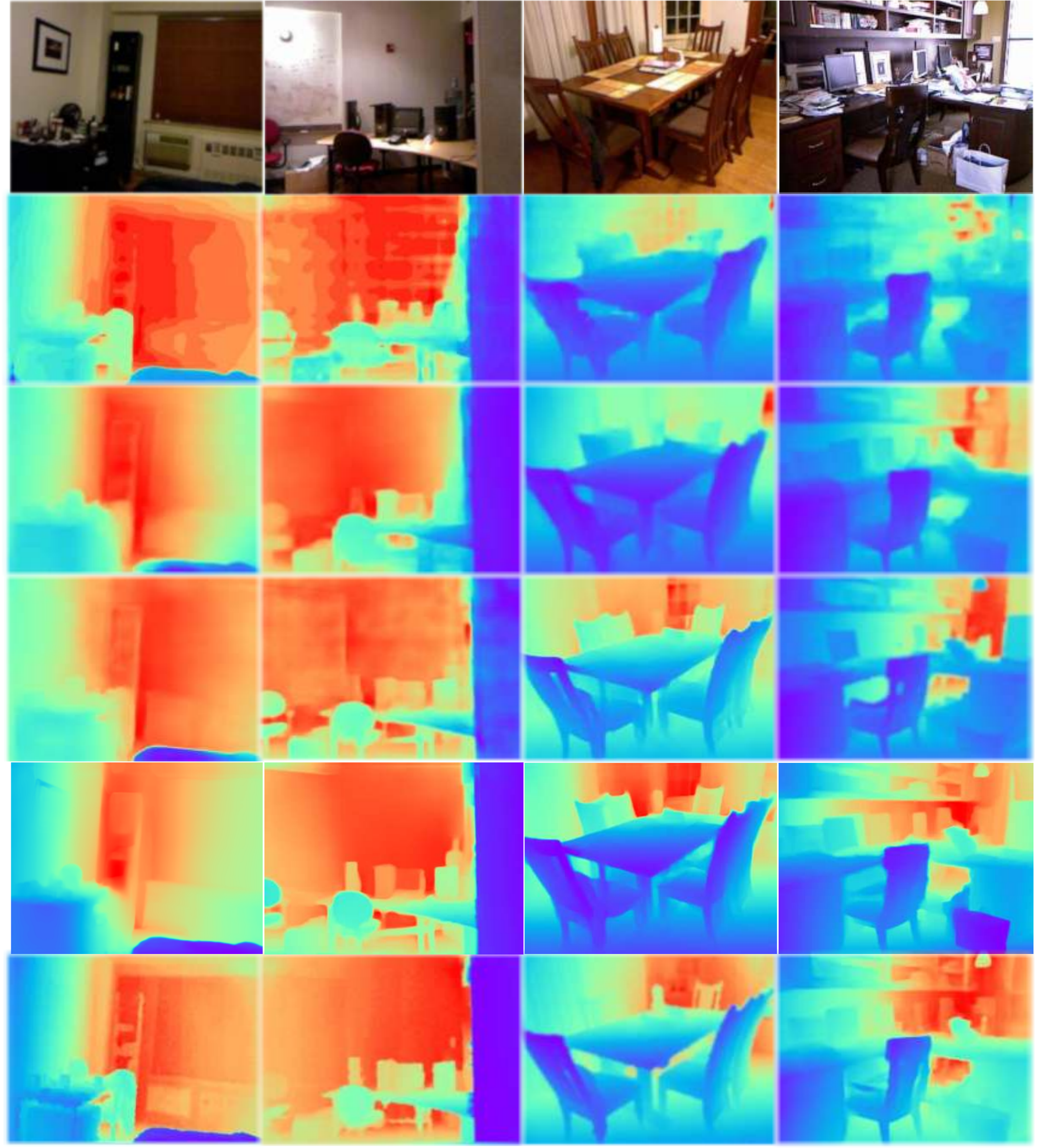}
     \end{minipage}
\end{minipage}
\caption{\textbf{Qualitative results on NYU Depth-v2.} We compare our method against SOTA methods using the same examples as~\cite{virtual:normal:depth}.}
\label{fig:nyu_results}
\vspace{-0.5cm}
\end{figure}

\begin{figure}[!t]
\begin{minipage}{0.99\columnwidth}
    \noindent\begin{minipage}{0.1\columnwidth}
        \footnotesize
        \begin{tabular}{l}
        \\ [-2em]
        \rotatebox{90}{\textbf{VNL}~\cite{virtual:normal:depth}}\\
        \\ [2em]
        \rotatebox{90}{\textbf{BTS}~\cite{local:planar:guidance}}\\
        \\ [2em]
        \rotatebox{90}{\textbf{Ours}}\\
        \\ [3em]
        \rotatebox{90}{\textbf{GT}}\\
        \\ 
        \end{tabular}
     \end{minipage}
     \noindent\begin{minipage}{0.89\columnwidth}
            \includegraphics[width=73mm]{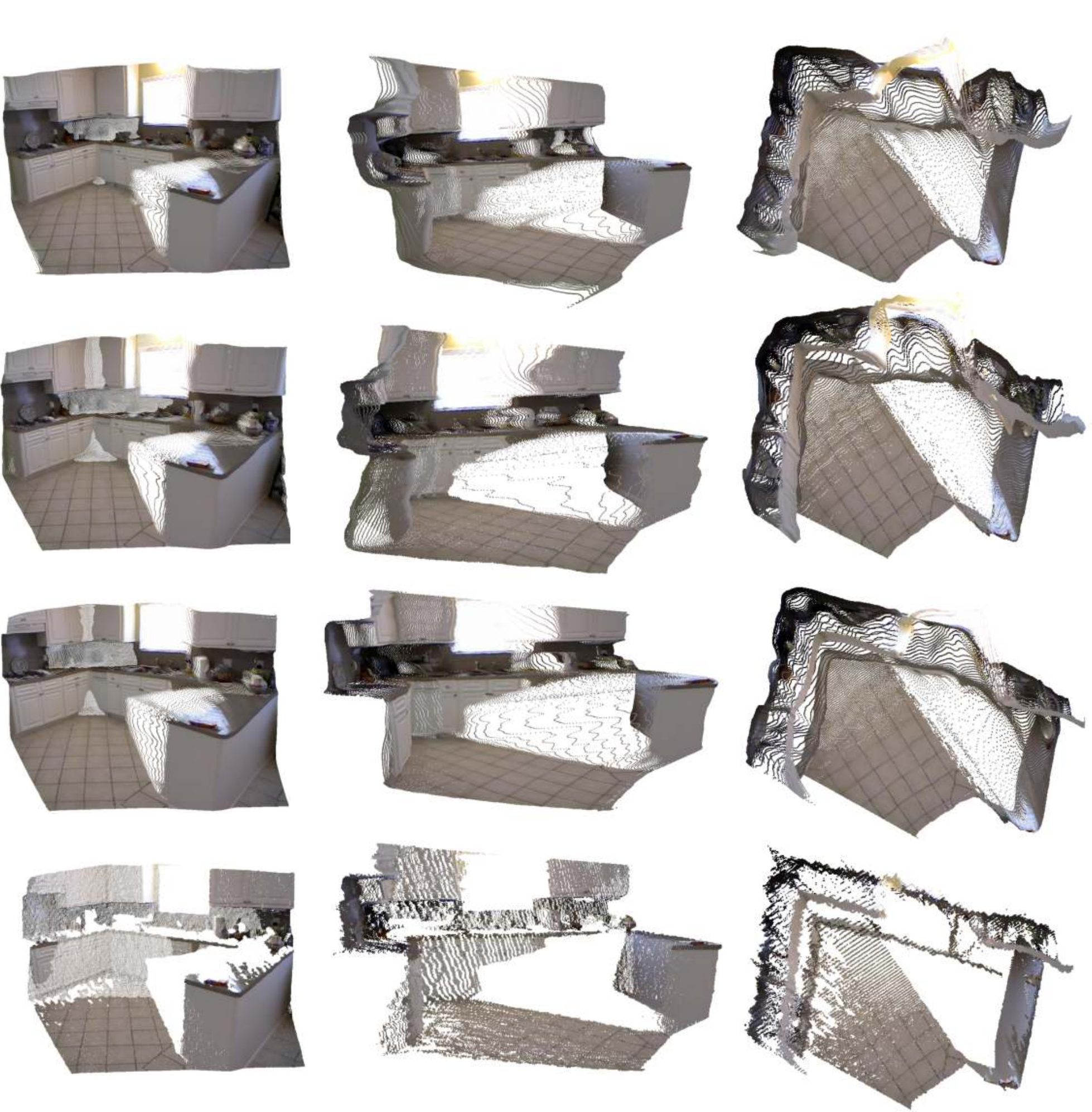}
            \begin{tabular}{c c c} \textbf{Front View} \hspace{0.5cm} & \textbf{Side View} \hspace{0.5cm} & \textbf{Top View }\\ \end{tabular}
     \end{minipage}
\end{minipage}
\caption{\textbf{Reconstruction example from NYU Depth-v2.} We compare the 3D reconstruction induced by our predicted depth to results using two SOTA depth estimation methods~\cite{virtual:normal:depth, local:planar:guidance}.}
\label{fig:pointclouds}
\vspace{-0.5cm}
\end{figure}

\section{Experiments}
\label{sec:experiments}

We structure this section as follows. We first discuss our experimental setup, i.e., datasets, evaluation metrics and implementation details for our method. We then compare our method to state-of-the-art approaches, followed by a thorough ablation study of our method.

\subsection{Experimental Setup}

In this section, we present the datasets used to evaluate our approach. The NYU Depth-v2 and KITTI datasets are used as the main sets for training and testing. Additionally, we use four more datasets for testing our method in a zero-shot transfer setup in order to evaluate its generalization potential. Evaluation on all six datasets is performed using standard depth evaluation metrics explained below.

\begin{table}
    \caption{\textbf{Datasets used in our experiments.} (*) uses mix of RGBD sensors. Sup.: supervision.}
    \label{table:datasets}
    \centering
    \setlength\tabcolsep{2pt}
    \footnotesize
    \begin{tabular*}{\linewidth}{l @{\extracolsep{\fill}} cccc}
    \toprule
    \textbf{Dataset} & \textbf{\# Training} & \textbf{\# Testing} &  \textbf{Sup.~Type} & \textbf{Scene Type}\\
    \midrule
    NYU Depth-v2~\cite{nyu} & 24,231 & 654 &  Kinect & Indoor\\
    KITTI~\cite{kitti} & 23,488 & 697 & LiDAR & Outdoor\\
    \midrule
    ScanNet~\cite{scannet} & - & 2167 & Kinect & Indoor\\
    SUN-RGBD~\cite{sunrgbd} & - & 5050 & Mixed* & Indoor\\
    DIODE Indoor~\cite{diode} & - & 325 & LiDAR & Indoor\\
    ETH-3D~\cite{eth3d} & - & 454 & LiDAR & Mixed\\
    \bottomrule
    \end{tabular*}
    \vspace{-0.6cm}
\end{table}

\PAR{NYU Depth-v2~\cite{nyu}.} The NYU Depth-v2 dataset consists of 464 indoor scenes of size 640$\times$480. These scenes are split into 249 scenes for training and 215 for testing. 
We use the official split provided by previous methods~\cite{local:planar:guidance} for training and the test set is based on~\cite{depth:multiscale:network}.

\PAR{KITTI~\cite{kitti}.} KITTI is an autonomous driving dataset consisting of 61 outdoor scenes of different types. We employ the standard depth estimation split proposed by Eigen \etal~\cite{depth:multiscale:network} and Garg \etal~\cite{unsupervised:cnn:single:view:depth}, for training and testing. We use 32 scenes for training and 29 scenes for testing.

\PAR{Datasets used for zero-shot testing.} To test the generalization of our P3Depth, we evaluate it on four datasets which are not seen during training: ScanNet, SUN-RGBD, DIODE Indoor and ETH-3D. The resolution of all images is reduced to 640$\times$480. Details are provided in Table~\ref{table:datasets}.

\PAR{Evaluation metrics.} We use the standard depth estimation metrics for evaluation. In particular, we use root mean square error (RMSE) and its log variant (RMSE\textit{log}), Log10 error, absolute (A.Rel) and squared (S.rel) mean relative error and the percentage of inlier pixels with $\delta$. The maximum depth for KITTI is set to 50m and 80m for the Garg and Eigen splits respectively. For NYU Depth-v2, the maximum depth is set to 10m as per the Eigen split. Zero-shot transfer is performed by using a model trained on NYU Depth-v2 without additional fine-tuning. The maximum depth for this task is set to 10m across all four test datasets. In all evaluations, the predicted depth is rescaled so that its median matches that of the ground truth, as per standard practice.

\begin{table}[!t]
    \caption{\textbf{Comparison of depth estimation methods on NYU Depth-v2~\cite{nyu} test set.} Comparison is performed on the Eigen split~\cite{depth:multiscale:network}.
    (nF) is number of frames, (*) indicates self-supervised methods and ($^\dagger$) denotes retrained results with train set from~\cite{local:planar:guidance}.}
    \resizebox{1\columnwidth}{!}{%
    \begin{tabular}{l|ccc|ccc}
    \toprule[1pt]
    \multicolumn{1}{c|}{\multirow{2}{*}{\textbf{Method}}} & \textbf{A.Rel} & \textbf{Log10} & \textbf{RMSE} & $\boldsymbol{\delta_{1}}$ & $\boldsymbol{\delta_{2}}$ & $\boldsymbol{\delta_{3}}$ \\
    & \multicolumn{3}{c|}{\textit{Lower is better}}         & \multicolumn{3}{c}{\textit{Higher is better}} \\ \hline
    \toprule[1pt]
    \multicolumn{7}{c}{\textbf{Plane detection based methods}} \\
    \hline
    PlaneNet~\cite{planenet:piecewise:planar:reconstruction} & 0.142 & 0.060 & 0.514 & 0.812 & 0.957 & 0.989 \\ 
    PlaneRCNN~\cite{planercnn:3d:plane:detection:reconstruction} & 0.124 & 0.077 & 0.644 & -- & -- & -- \\ 
    Yu \etal~\cite{piecewise:planar:3d:reconstruction:associative:embedding}    &  0.134 &  0.057  &  0.503 &  0.827 &  0.963 & 0.990   \\
    $\text{P}^{2}$ Net (5F)*~\cite{p2net:unsupervised:indoor:depth} & 0.147 & 0.062 & 0.553 & 0.801 & 0.951 & 0.987 \\
    StruMonoNet~\cite{StruMonoNet} & 0.107 & 0.046 & 0.392 & 0.887 & \underline{0.980} & 0.995\\
    \toprule[1pt]
    \multicolumn{7}{c}{\textbf{Other monocular depth estimation methods}} \\
    \hline
    Saxena \etal.~\cite{depth:make3d}  & 0.349  & -    & 1.214   & 0.447  & 0.745  & 0.897\\
    Karsch \etal.~\cite{depth:transfer:pami:2014}   & 0.349  & 0.131  & 1.21   & -  & -  & -          \\
    Liu \etal.~\cite{depth:discrete:cvpr:2014}     & 0.335  & 0.127   & 1.06  & -  & -  & -          \\
    Ladicky \etal.~\cite{depth:pulling:cvpr:2014}  & -   & -  & -    & 0.542  & 0.829  & 0.941     \\
    Li \etal.~\cite{depth:hierar:cvpr:2015}   & 0.232  & 0.094  & 0.821   & 0.621 & 0.886  & 0.968 \\
    Wang \etal.~\cite{depth:semantic:unified:cvpr:2015}  & 0.220  & 0.094  & 0.745  & 0.605  & 0.890 & 0.970 \\
    Liu \etal.~\cite{depth:neural:fields:pami:2016}   & 0.213  & 0.087  & 0.759  & 0.650  & 0.906 & 0.974  \\
    Roy \etal.~\cite{depth:neural:forest:cvpr:2016}   & 0.187 & 0.078  & 0.744  & -  & -   & -          \\
    AdaBins$^\dagger$~\cite{adabins} & 0.178 & 0.078 & 0.595 & 0.698 & 0.937 & 0.988 \\
    Eigen \etal.~\cite{depth:multiscale:network}  & 0.158  & -     & 0.641 & 0.769  & 0.950  & 0.988 \\
    Chakrabarti~\cite{depth:harmonizing:nips:2016}      & 0.149  & - & 0.620  & 0.806  & 0.958  & 0.987 \\
    Li \etal.~\cite{depth:two:stream:iccv:2017}   & 0.143   & 0.063  & 0.635  & 0.788  & 0.958 & 0.991    \\
    Laina \etal.~\cite{depth:frcn}   & 0.127  & 0.055    & 0.573  & 0.811   & 0.953  & 0.988      \\
    Fu \etal~\cite{depth:ordinal:cvpr:2018}   & 0.115  & 0.051  & 0.509   & 0.828  & 0.965  & 0.992  \\ 
    Yin \etal~\cite{virtual:normal:depth} & 0.108 &  0.048 & 0.416 & 0.875 & 0.976 & 0.994\\
    Huynh \etal~\cite{depth:attention:volume} & 0.108 & - & 0.412 & 0.882 & 0.980 & \underline{0.996}\\
    Lee \etal~\cite{local:planar:guidance} & 0.110 & 0.047 & 0.392 & 0.885 & 0.978 & 0.994\\
    Long \etal~\cite{adaptive:surface:normal:constraint:depth} & \textbf{0.101} & \underline{0.044} & 0.377 & 0.890 & \underline{0.982} & \underline{0.996} \\
    Ranftl \etal~\cite{vision:transformers:depth:iccv:2021} & 0.110 & 0.045 & \underline{0.357} & \textbf{0.904} & \textbf{0.988} & \textbf{0.998} \\
    \hline \hline
    Ours & \underline{0.104}  & \textbf{0.043} & \textbf{0.356} & \underline{0.898}  & 0.981 & \underline{0.996} \\
    \toprule[1pt]
    \end{tabular}}
    \label{table:NYUD-V2}
    \vspace{-0.70cm}
\end{table}

\PAR{Implementation details.} Our network includes two heads. The first head outputs four channels, one for each plane coefficient. The second head outputs three channels: one for each coordinate of the offsets and one for confidence. These heads are fed by a ResNet101 encoder~\cite{resnet} initialized with pre-trained ImageNet~\cite{imagenet} weights. This initialization is important to achieve competitive results as in~\cite{local:planar:guidance,depth:attention:volume,virtual:normal:depth,depth:ordinal:cvpr:2018}. The decoders, inspired from ~\cite{decoders}, are initialized with weights drawn from a normal distribution with $\sigma$= 0.01. The plane coefficient decoder is additionally equipped with a guidance module. See  for details.
The offset vectors are restricted via a tanh layer to have a maximum length of $\tau$ in normalized image coordinates. We set $\tau$ to 0.1 by default and apply two steps of cascaded refinement to the offsets. The confidence map is predicted through a sigmoid layer. For all experiments, we use a batch size of 8 and employ the Adam optimizer~\cite{adam} with a learning rate of $10^{-4}$ and a weight decay of $10^{-4}$. We train our network for 25 epochs, although the model starts producing decent predictions from epoch 5. The learning rate is reduced every 5 epochs by a factor of 10 using a step scheduler. The training images are resized similarly to~\cite{local:planar:guidance}. For all direct depth losses, we use the loss formulation from~\cite{depth:multiscale:network}. The loss weights $\lambda$ and $\mu$ are set to 0.5. In addition, the mean plane loss is applied using the final depth prediction $D_f$. The offset head performs better with dense supervision. Hence, $D_s$ is supervised using completed $D^*$. To complete $D^*$, the depth inpainting method from~\cite{nyu} is used. The inpainted ground truth is also used for computing the mean plane loss to provide stability to the SVD algorithm for least squares. We set the patch size to 32 and $K$ in \eqref{eq:MPL} is set indirectly by the patch size and the image size.
The data augmentation techniques from~\cite{local:planar:guidance} are used.

\begin{table}[!t]
  \caption{\textbf{Comparison of depth estimation methods on KITTI~\cite{kitti}.} Comparison is performed on the Eigen test split.}
  \label{table:exp:kitti}
	\centering
	\resizebox{1\columnwidth}{!}{%
		\begin{tabular}{l|cccc|ccc}
			\hline
			\multicolumn{1}{c|}{\multirow{2}{*}{\textbf{Method}}} & \textbf{A.Rel} & \textbf{S.Rel} & \textbf{RMSE} & \textbf{RMSE\textit{log}} & $\boldsymbol{\delta_{1}}$ & $\boldsymbol{\delta_{2}}$ & $\boldsymbol{\delta_{3}}$ \\ & \multicolumn{4}{c|}{\textit{Lower is better}} & \multicolumn{3}{c}{\textit{Higher is better}} \\ \hline
			\multicolumn{8}{c}{\textbf{Garg split~\cite{unsupervised:cnn:single:view:depth} cap: 50m}} \\ \hline
			Garg \etal~\cite{unsupervised:cnn:single:view:depth}            & 0.169 & 1.080 & 5.104 & 0.273 & 0.740 & 0.904 & 0.962 \\
			Godard \etal~\cite{monodepth} & 0.108 & 0.657 & 3.729 & 0.194 & 0.873 & 0.954 & 0.979 \\
			Kuznietsov~\cite{semi:supervised:learning:monocular:depth}        & 0.108 & 0.595 & 3.518 & 0.179 & 0.875 & 0.964 & 0.988 \\
			Gan \etal~\cite{depth:pooling:eccv:2018}                & 0.094 & 0.552 & 3.133 & 0.165 & 0.898 & 0.967 & 0.986 \\
			Fu \etal~\cite{depth:ordinal:cvpr:2018}          & 0.071 & 0.268 & 2.271 & 0.116 & 0.936 & 0.985 & 0.995 \\ 
			AdaBins~\cite{adabins} & 0.058 & 0.19 & 2.36 & 0.088 & \underline{0.964} & \underline{0.995} & \textbf{0.999} \\ 
			Lee \etal~\cite{local:planar:guidance}             & \underline{0.056} & \underline{0.169} & \underline{1.925} & \underline{0.087} & \underline{0.964} & 0.994 & \textbf{0.999} \\
			\hline \hline 
             Ours             & \textbf{0.055} & \textbf{0.130} & \textbf{1.651} & \textbf{0.081} & \textbf{0.974} & \textbf{0.997} & \textbf{0.999} \\
			\toprule[1pt]
			\multicolumn{8}{c}{\textbf{Eigen split~\cite{depth:multiscale:network} cap: 80m}} \\ \hline
			Saxena \etal~\cite{depth:make3d}                  & 0.280 & 3.012 & 8.734 & 0.361 & 0.601 & 0.820 & 0.926  \\
			Eigen \etal~\cite{depth:multiscale:network}       & 0.203 & 1.548 & 6.307 & 0.282 & 0.702 & 0.898 & 0.967 \\
			Liu \etal~\cite{depth:convolutional:neural:fields} & 0.201 & 1.584 & 6.471 & 0.273 & 0.680 & 0.898 & 0.967 \\
			Godard \etal~\cite{monodepth} & 0.114 & 0.898 & 4.935 & 0.206 & 0.861 & 0.949 & 0.976 \\
			Kuznietsov~\cite{semi:supervised:learning:monocular:depth}        & 0.113 & 0.741 & 4.621 & 0.189 & 0.862 & 0.960 & 0.986 \\
			Gan \etal~\cite{depth:pooling:eccv:2018}                 & 0.098 & 0.666 & 3.933 & 0.173 & 0.890 & 0.964 & 0.985 \\
			Fu \etal~\cite{depth:ordinal:cvpr:2018}           & 0.072 & 0.307 & \underline{2.727} & 0.120 & 0.932 & 0.984 & 0.994 \\ 
			Yin \etal~\cite{virtual:normal:depth}             & 0.072 &   -   & 3.258 & 0.117 & 0.938 & 0.990 & \underline{0.998} \\
			Lee \etal~\cite{local:planar:guidance}             & \textbf{0.059} & \textbf{0.245} & 2.756 & \underline{0.096} & \underline{0.956} & \underline{0.993} & \underline{0.998} \\ 
			AdaBins~\cite{adabins} & 0.067 & 0.278 & 2.96 & 0.103 & 0.949 & 0.992 & \underline{0.998} \\
			Ranftl \etal~\cite{vision:transformers:depth:iccv:2021} & \underline{0.062} & - & \textbf{2.573} & \textbf{0.092} & \textbf{0.959} & \textbf{0.995} & \textbf{0.999} \\
			\hline \hline 
             Ours   & 0.071 & \underline{0.270} & 2.842 & 0.103 & 0.953 & \underline{0.993} & \underline{0.998} \\
			\toprule[1pt]
		\end{tabular} }
	\vspace{-0.7cm}
\end{table}

\subsection{Comparison with State of the Art}

\textbf{NYU Depth-v2:}
The results on NYU Depth-v2, which is the major indoor depth benchmark, are presented in Table~\ref{table:NYUD-V2}. We set the new state of the art on NYU Depth-v2, outperforming prior state-of-the-art (SOTA) methods across all six standard metrics. We achieve a superior relative performance gain of 9.18\% in RMSE and 3.7\% in A.Rel, while also improving $\delta_{1}$ by 1.1\%. This improvement in performance indicates that without using ground-truth planes as supervision, P3Depth learns an implicit representation of the planes which can benefit the overall depth estimation capability of the network.

Qualitative results on NYU Depth-v2 support the above findings. In Fig.~\ref{fig:nyu_results}, we show the high-quality predictions generated by our method in comparison with SOTA methods using the same examples as in~\cite{virtual:normal:depth}. It can be clearly observed that the surfaces which fit our piece-wise planar assumption, such as table, cupboard, and even smaller objects, e.g.\ computer screens, photo frames etc.\ have consistent predictions with sharp details in comparison with the SOTA methods. Overall, our method generates superior visual results. In some cases, especially w.r.t.\ the metric scale of the scene, results from~\cite{local:planar:guidance} are comparable to ours. Our method excels especially on man-made regular structures of the indoor scenes. What is more, the predicted depth maps produce 3D reconstructions which are consistent with ground-truth point clouds and preserve the structure of the scene better than competing methods, as shown in Fig.~\ref{fig:pointclouds}.

\begin{figure}[t]
\includegraphics[width=\linewidth]{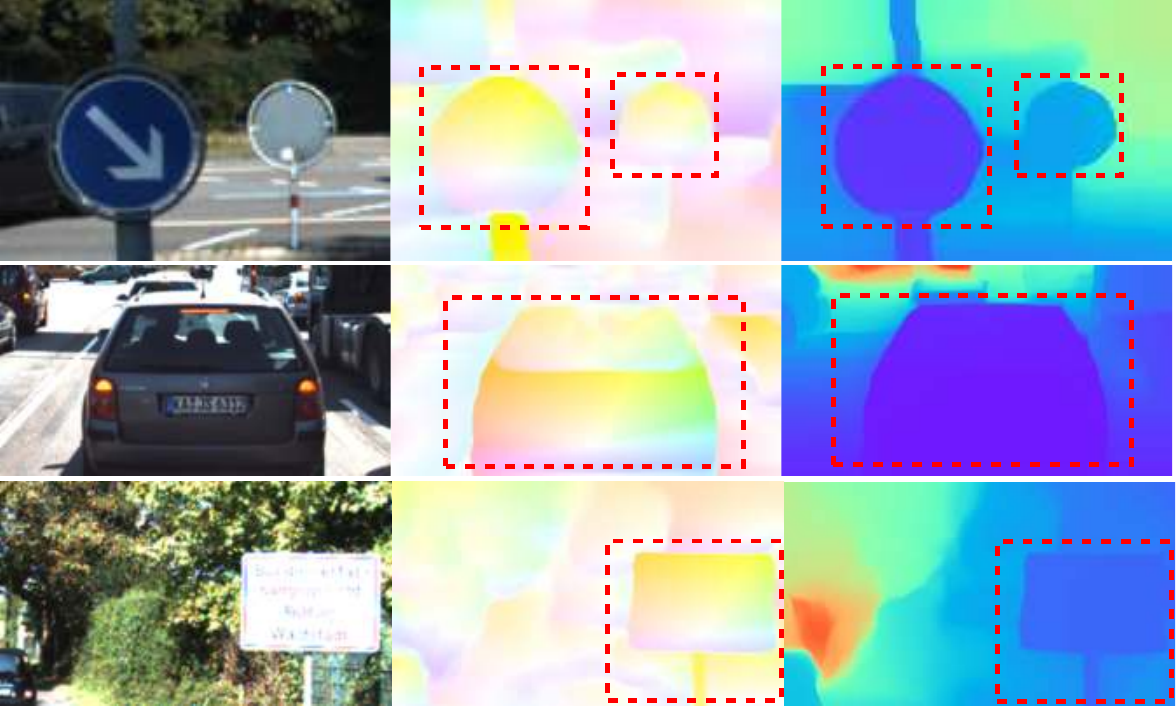}
\caption{\textbf{Qualitative results on KITTI.} We present the predicted offset vector fields (middle) and the depth estimates (right). }
\label{fig:kitti_results}
\vspace{-0.6cm}
\end{figure}

\textbf{KITTI:} The results on KITTI in Table~\ref{table:exp:kitti} suggest that our method is fully applicable to outdoor datasets. In particular, we surpass prior state of the art on the Garg split (with maximum range of 50m) in all metrics by a significant margin. More specifically, we improve RMSE by 14.2\% and $\delta_1$ by 1.0\%. This proves that our method takes advantage of regular structures in outdoor scenes to improve depth predictions. Moreover, our method is comparable to state of the art on the Eigen split~\cite{depth:multiscale:network}, where the maximum range is 80m. The reason why our ranking is slightly lower on the Eigen split is that distant parts of the scene get projected to smaller regions and thus the extent of the respective smooth pieces of the depth map is also smaller, making it more difficult to predict a correct offset. Additionally, for the results on the KITTI benchmark suite, please refer to the appendix~\ref{appendix:results}. Overall, the method is able to handle planar objects quite well even in variable lighting conditions as shown in Fig.~\ref{fig:kitti_results}. Although the sign board in the bottom image is brightly lit and the car in the middle image is badly lit, the method is able to detect the regularity of the object surface and predict consistent depth. The top image of Fig.~\ref{fig:kitti_results} demonstrates a limitation of our method. In particular, the road segments on either side of the pole of the traffic sign on the right get mapped to significantly different depth values because they are disconnected and thus do not interact in terms of plane coefficients, even though they belong to the same 3D plane. Additionally, the planar specular glass surface of the car in middle image is predicted incorrectly. This is due to the shortcomings of the sensor used to measure depth. The erroneous ground truth does not allow the network to learn the depth or the offset vector field in these regions.

\textbf{Zero-shot experiments:} In Table~\ref{table:exp:zeroshot}, we prove the generalization ability of our method in a zero-shot setting where the test domains have not been seen during training. We achieve the best performance on the ScanNet~\cite{scannet} and SUN-RGBD~\cite{sunrgbd} indoor datasets in all metrics. On DIODE Indoor and ETH-3D, \cite{virtual:normal:depth} performs best in terms of A.Rel, but we are by far the best in terms of both RMSE and $\delta_{1}$. This comparison shows that even when our method is trained only on an indoor dataset as NYU Depth-v2, it works well on a variety of datasets with different types of scenes.

\subsection{Ablation Studies}

We study the importance of the components of our method by ablating them in Table~\ref{tab:ablation}. We observe that in a standalone setting, directly predicting depth is better than predicting planar coefficients. However, once we insert the second head which predicts offset vectors, a substantial benefit is obtained by using the plane coefficient representation compared to directly predicting depth. This demonstrates that the network learns to make effective use of local planar information at seed pixels to improve depth, thanks to the plane coefficient representation.
Moreover, adding our guidance module provides a slight improvement. The ablation also verifies the utility of cascaded refinement of offsets, which yields better results than simply using a higher maximum offset length. Finally, adding our mean plane loss on top of plane coefficients, offset vectors and cascaded offset refinement yields SOTA results on NYU Depth-v2.

\begin{table}
  \caption{\textbf{Comparison of SOTA methods on generalized learning of metric depth.} All methods are trained on NYU Depth-v2 and tested on four other datasets without fine-tuning.}
  \label{table:exp:zeroshot}
  \centering
  \footnotesize
  \resizebox{1\columnwidth}{!}{%
  \begin{tabular}{ l | l | c c c }
  \toprule[1pt]
  \textbf{Dataset} & Metric & VNL~\cite{virtual:normal:depth} & BTS~\cite{local:planar:guidance} & Ours \\
  \hline
  \multirow{3}{*}{ScanNet~\cite{scannet}} & A.Rel $\downarrow$ & 0.227 & 0.255 & \textbf{0.223} \\
  & RMSE $\downarrow$ & 0.563 & 0.615 & \textbf{0.538} \\
  & $\delta_{1}$ $\uparrow$ & 0.544 & 0.472 & \textbf{0.551} \\
  \hline
    \multirow{3}{*}{SUN-RGBD~\cite{sunrgbd}} & A.Rel $\downarrow$ & 0.317 & 0.317 & \textbf{0.307} \\
  & RMSE $\downarrow$ & 0.449 & 0.461 & \textbf{0.431} \\
  & $\delta_{1}$ $\uparrow$ & 0.793 & 0.794 & \textbf{0.797} \\
  \hline
    \multirow{3}{*}{Diode Indoor~\cite{diode}} & A.Rel $\downarrow$ & \textbf{0.291} & 0.310 & 0.373 \\
  & RMSE $\downarrow$ & 0.890 & 0.981 & \textbf{0.784} \\
  & $\delta_{1}$ $\uparrow$ & 0.635 & 0.559 & \textbf{0.639} \\
  \hline
    \multirow{3}{*}{ETH-3D~\cite{eth3d}} & A.Rel $\downarrow$ & \textbf{0.331} & 0.366 & 0.343 \\
  & RMSE $\downarrow$ & 1.649  & 1.840 & \textbf{1.637} \\
  & $\delta_{1}$ $\uparrow$ & 0.462 & 0.398 & \textbf{0.468} \\
  \bottomrule
  \end{tabular}
  }
  \vspace{-0.15cm}
\end{table}

\begin{table}
    \footnotesize
    \centering
    \caption{\textbf{Ablation study of components of our method.} $D$: directly predicting depth, $\mathbf{C}$: predicting plane coefficients, ``Guid.'': guidance module for plane coefficient decoder, ``OV'': offset vectors, ``Ref.'': cascaded refinement of offsets, ``MPL'': mean plane loss, ``+'': offset length is restricted to $\tau$=0.3 instead of $\tau$=0.1.}
    \label{tab:ablation}
    \resizebox{1\columnwidth}{!}{%
        \begin{tabular}{lllll|cc|c}
        \toprule[1pt]
        \footnotesize \textbf{Pred.}
        & \footnotesize \textbf{Guid.}
        & \footnotesize \textbf{OV}
        & \footnotesize \textbf{Ref.}
        & \footnotesize \textbf{MPL}
        & \textbf{A.Rel} $\downarrow$
        & \textbf{RMSE} $\downarrow$
        & $\boldsymbol{\delta_{1}}$ $\uparrow$\\
        \hline
        $D$ & & & & & 0.142  & 0.458 & 0.821  \\
        $\mathbf{C}$ & & & & & 0.144 & 0.487 & 0.811 \\
        \hline
        $\mathbf{C}$ & \checkmark & & & & 0.142  & 0.458 & 0.824  \\
        \hline
        $D$ &  & \checkmark & &  & 0.140  & 0.453 & 0.824  \\
        $\mathbf{C}$ &  & \checkmark & &  & 0.116 & 0.390 & 0.877  \\
        \hline
        $\mathbf{C}$ &  &  &  & \checkmark & 0.118 & 0.395 & 0.872  \\
        \hline
        $\mathbf{C}$ & \checkmark & \checkmark &  &  & 0.115  & 0.384 & 0.879  \\
        $\mathbf{C}$ & \checkmark & \checkmark+ &  &  & 0.116  & 0.390 & 0.879  \\
        \hline
        $D$ &  & \checkmark & \checkmark &  & 0.134 & 0.440 & 0.839 \\
        $\mathbf{C}$ &  & \checkmark & \checkmark &  & 0.113 & 0.378 & 0.884  \\
        $\mathbf{C}$ &  & \checkmark & \checkmark & \checkmark & 0.109 & 0.370 & 0.890  \\
        \hline
        $\mathbf{C}$ & \checkmark & \checkmark &  \checkmark &  & 0.109  & 0.373 & 0.889  \\
        $\mathbf{C}$ & \checkmark & \checkmark & \checkmark & \checkmark & \textbf{0.104}  & \textbf{0.356} & \textbf{0.898}  \\
        \bottomrule[1pt]
        \end{tabular}
    }
    \vspace{-0.5cm}
\end{table}

\section{Conclusion}

We have presented a supervised method for monocular depth estimation which leverages local planar information in the 3D scene in order to predict consistent depth values across smooth parts of the scene. The method uses a plane coefficient representation for depth, which enables to share information from seed locations and improve the predicted depth. We implicitly learn to predict the offsets to these seed locations and to weigh the depth obtained from them adaptively according to accuracy. We empirically validated our method on the major indoor and outdoor benchmarks for monocular depth estimation and set the new state of the art among supervised approaches, which shows the potential of well-selected geometric priors for depth estimation.

\PAR{Acknowledgements.}
This work is funded by Toyota Motor Europe via the research project TRACE-Z\"urich. We thank Guolei Sun for sharing his GPU quota.

{\small
\bibliographystyle{ieee_fullname}
\bibliography{refs}
}

\clearpage
\appendix

\section{Network Architecture}
\label{appendix:network}
\PAR{Encoder:}
Following the recent works~\cite{deep:ordinal:regression:network, local:planar:guidance}, we use a ResNet101~\cite{resnet} as the encoder for the image. Each ResNet block consists of series of convolution operations with stride of 2 and pooling operations. The receptive field of the convolution is increased by decreasing resolution of the feature maps. This helps to capture more contextual information while compromising the feature map resolution. The final size of the feature map is usually 1/32 of the input image. The original ResNet is designed for the image classification task. To utilize it for a per-pixel prediction task, we remove the last 3 layers, i.e. pooling layer, fully-connected layer and the softmax layer. The ResNet encoder can be divided into 4 different blocks. Each block generates feature maps of different resolution (scales). These feature maps from different scales can be used as skip connections, i.e. fused with decoder outputs to integrate different level of semantic information. The output of the last encoder block is fetched to both decoder heads. Both decoder heads also receive the skip connection information. 

\PAR{Decoder:} We base our decoder on~\cite{decoders} following~\cite{structure:guided:ranking:loss:depth}. We replace all ReLU operations with ELU~\cite{elu} nonlinearities. The decoder is assembled from three modules: 1) \textit{Feature fusion modules:} For each of these modules, residual convolution block is used to transform the skip connection feature map from the ResNet encoder. The output of the residual convolution block is fused with output of last feature fusion block using summation operation. Finally, the feature maps are upsampled to match the resolution of next layers input. 2) \textit{Residual convolution modules:} This module is a series of two units of ELU and $3\times3$ convolution operations to merge the output of a previous decoder feature map output with a previous feature fusion module output 3) \textit{Adaptive output module:} This is applied at the last stage to get the final output. It consist of two $3\times3$ convolution operation followed by up-sampling.\\ 
\textbf{Plane coefficient decoder:} The last layer of this decoder head is modified to output 4-channels for each planar coefficient instead of single channel depth.\\ 
\textbf{Offset Vector field decoder:}  The last layer of this decoder head is modified to output 3-channels, i.e. two channels for the offset vector field and one for the confidence. The offset vector field is restricted by $\tanh$ layers and the confidence is generated through a sigmoid layer. \\
\textbf{Plane coefficient guidance:} This module is loosely based on~\cite{local:planar:guidance}. The output of each decoder block is passed through the Plane coefficient guidance module to generate 4 channels of plane coefficients. The output size of the guidance module is up-sampled to match the input size of last decoder layer. 
At the end, these plane coefficients from each scale are converted into depth. All these depth maps are concatenated with feature map of the previous decoder layer passed to the last decoder layer. 

\begin{figure}[t]
\includegraphics[width=\linewidth]{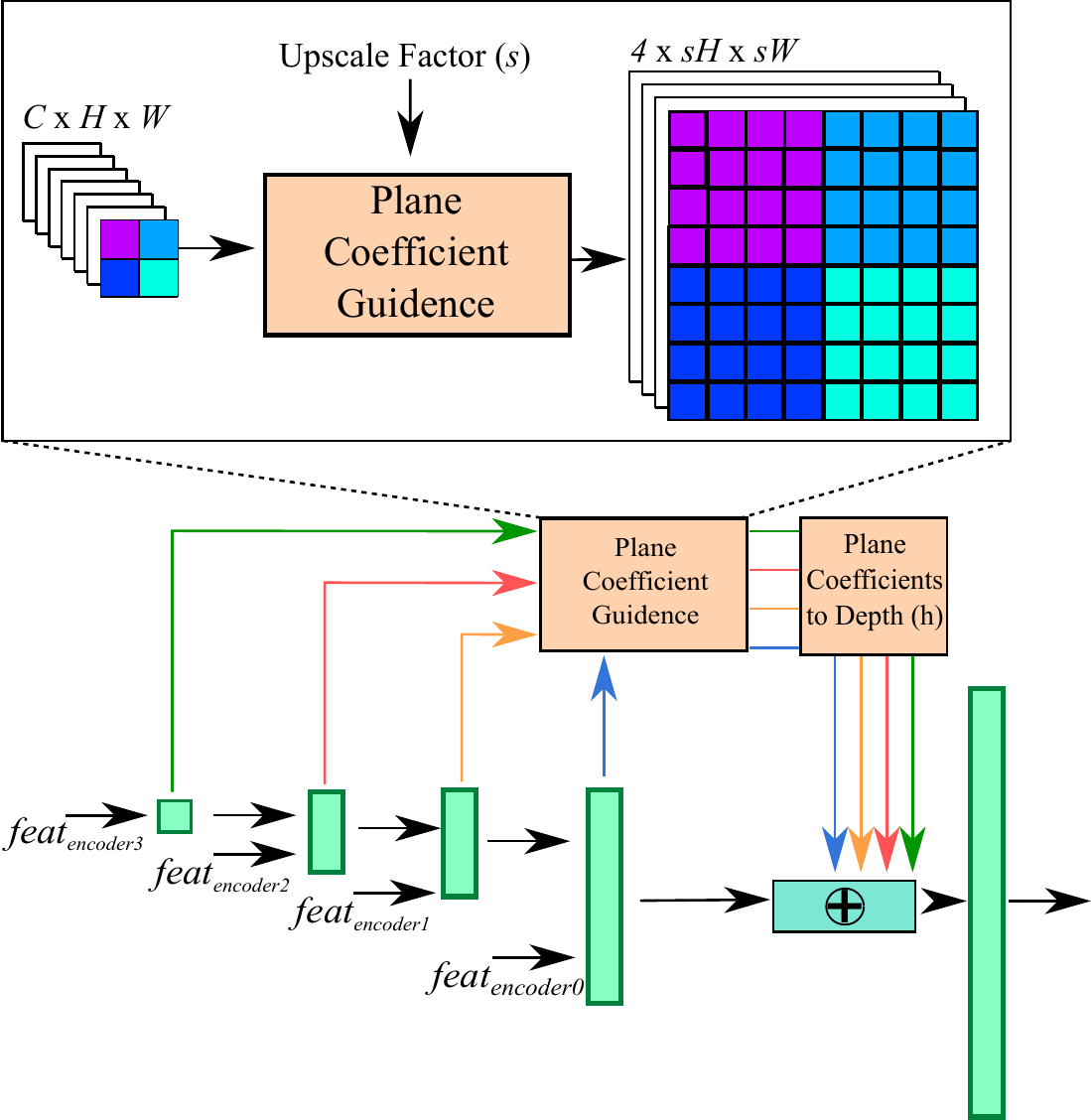}
\caption{\textbf{Plane coefficient guidance module.}}
\label{fig:guidance_module}
\end{figure}

\section{Additional Results}
\label{appendix:results}
\PAR{KITTI Benchmark~\cite{kitti}:}
In this section we present the results of KITTI Benchmark server evaluation. Note that we train our model only on the KITTI Eigen split~\cite{depth:multiscale:network} training data. It can be seen in Table~\ref{table:errors_kitti_benchmark} that our results are on a par with SOTA methods and superior than the baseline. However, ~\cite{local:planar:guidance} performs better on this test set. In comparison with ~\cite{virtual:normal:depth}, we have a better absolute relative error and our performance is comparable to~\cite{virtual:normal:depth} in all other metrics. The drop in overall performance is expected considering the design of our method. Our method is specially designed to identify planar regions in the scene, to improve the depth quality. So, as the depth of the scene increases, the projections of distant parts of the scene get smaller. This causes difficulties in predicting offset vector field in these regions. We have already seen that our method produced the SOTA results on the Garg split~\cite{unsupervised:cnn:single:view:depth}, in which the maximum depth value is 50m. Due to the aforementioned reason, when tested on Eigen split~\cite{depth:multiscale:network} with max depth of 80m, we observe degradation in the performance. The KITTI Benchmark extends beyond that with 80m+ distances, thus affecting our results due to similar reasons.     

\begin{table}[H]
 \caption{\textbf{Results of KITTI Evaluation Server.}}
\resizebox{1\columnwidth}{!}{%
\begin{tabular}{r |cccc}
\toprule[1pt]
\multicolumn{1}{c|}{\multirow{1}{*}{\textbf{Method}}} & \textbf{SILog} & \textbf{sqErrorRel} & \textbf{absErrorRel} & \textbf{iRMSE} \\ \hline
Official Baseline & 18.19 & 7.32 & 14.24 & 18.50 \\
VNL~\cite{virtual:normal:depth} & \underline{12.65}  & \underline{2.46}& 10.15 & \underline{13.02} \\
BTS~\cite{local:planar:guidance} & \textbf{11.67} & \textbf{2.21} & \textbf{9.04} & \textbf{12.23} \\
\hline
Ours & 12.82  & 2.53 & \underline{9.92} & 13.71 \\
\toprule[1pt]
\end{tabular}}
\label{table:errors_kitti_benchmark}
\end{table}

\PAR{Qualitative Results:}
Here, we present additional qualitative results on both KITTI~\cite{kitti} and NYU Depth-v2~\cite{nyu} datasets. We start with some examples from the KITTI dataset. We present some of the best cases along with the failure cases on this dataset. Additionally, we provide visualizations of the predicted depth maps and offset vector fields on NYU Depth-v2. Finally, we use the predicted depth maps to reconstruct the scenes and demonstrate quality in 3D. We observe that the predicted depth maps produce 3D reconstructions which are consistent with ground-truth point clouds and preserve the structure of the scene.

\begin{figure*}
\centering
\begin{minipage}{0.99\textwidth}
   \includegraphics[width=1\linewidth]{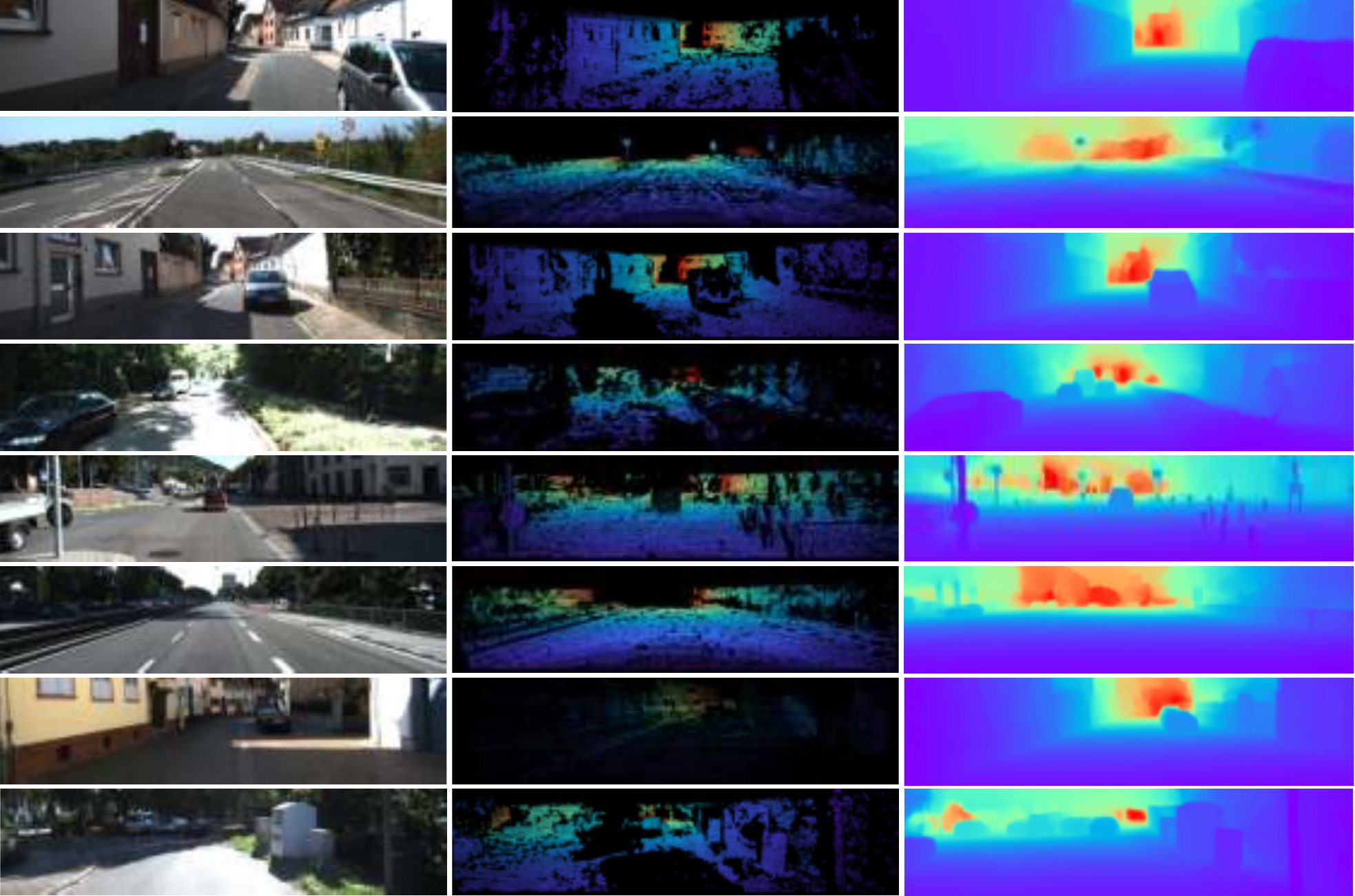}
    \begin{tabular}{c c c} \hspace{2cm} \textbf{Image} \hspace{3.5cm} & \textbf{Ground-truth Depth} \hspace{2.5cm} & \textbf{Predicted Depth}\\ \end{tabular}
   \caption{\textbf{Visualization of predictions on KITTI dataset.}}
   \label{fig:kitti_best_cases} 
   \vspace{2cm}
\end{minipage}

\begin{minipage}{0.99\textwidth}
   \includegraphics[width=1\linewidth]{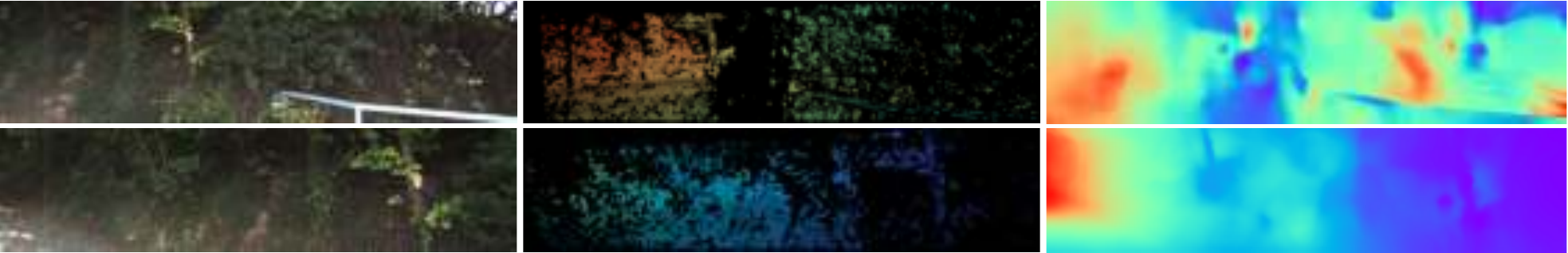}
       \begin{tabular}{c l l} \hspace{2cm} \textbf{Image} \hspace{3.5cm} & \textbf{Ground-truth Depth} \hspace{2.5cm} & \textbf{Predicted Depth}\\ \end{tabular}
   \caption{\textbf{Visualization of some failure cases on KITTI dataset.}}
   \label{fig:kitti_failure_cases}
\end{minipage}
\end{figure*}

\begin{figure*}[!t]
\centering
\includegraphics[width=0.8\linewidth]{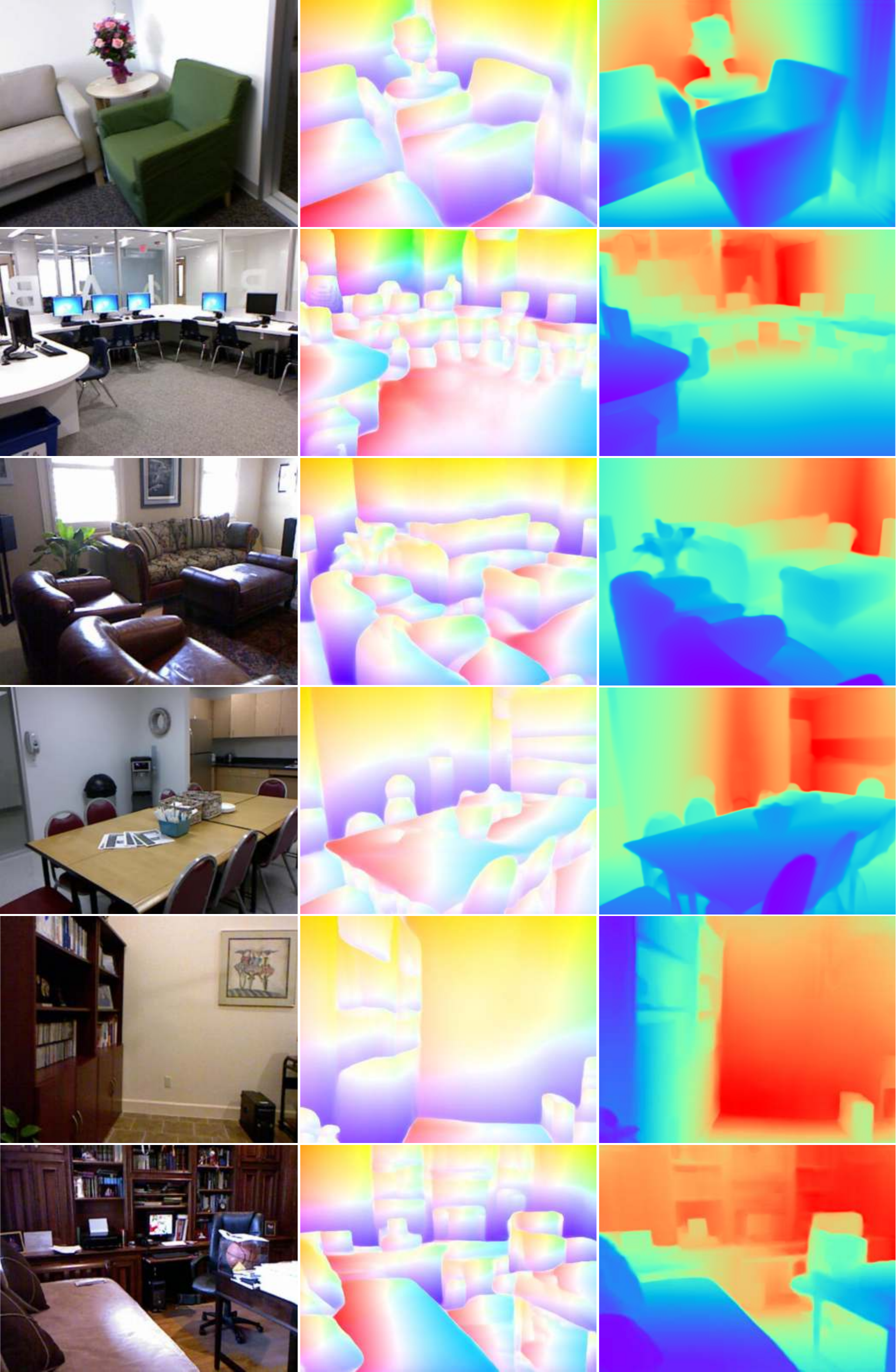}
\begin{tabular}{cccccc} \hspace{1cm}  &  \textbf{Image}  &  \hspace{2cm}  &\hspace{-0.5cm} \textbf{Predicted off. vec. field}  &   \hspace{1cm} \textbf{Predicted Depth} & \\ \end{tabular}
\caption{\textbf{Visualization of predictions on NYU Depth-v2.}}
\label{fig:nyu_best_cases}
\end{figure*}

\begin{figure*}
\centering
\begin{minipage}{0.99\textwidth}
\centering
   \includegraphics[width=0.99\linewidth]{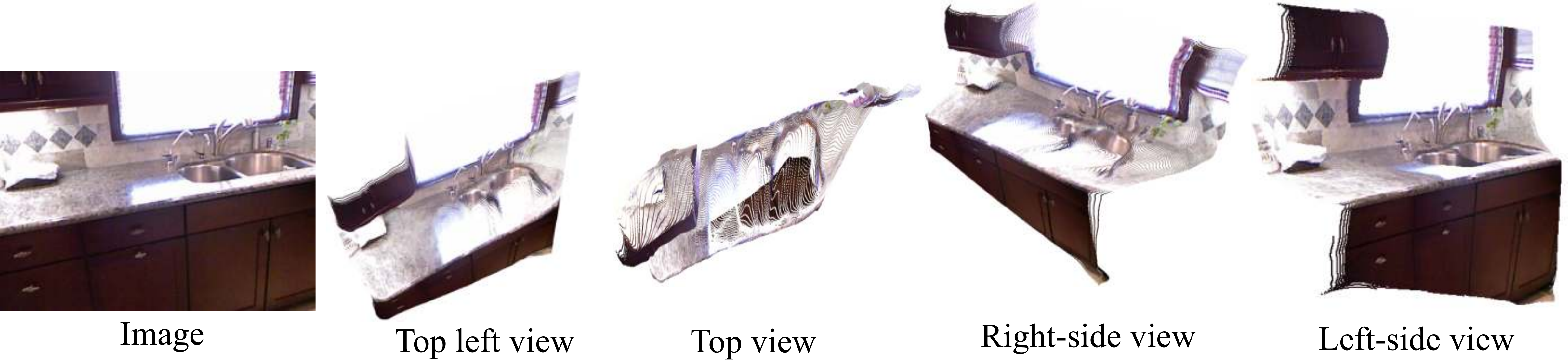}
   \vspace{2cm}
\end{minipage}

\begin{minipage}{0.99\textwidth}
\centering
   \includegraphics[width=0.99\linewidth]{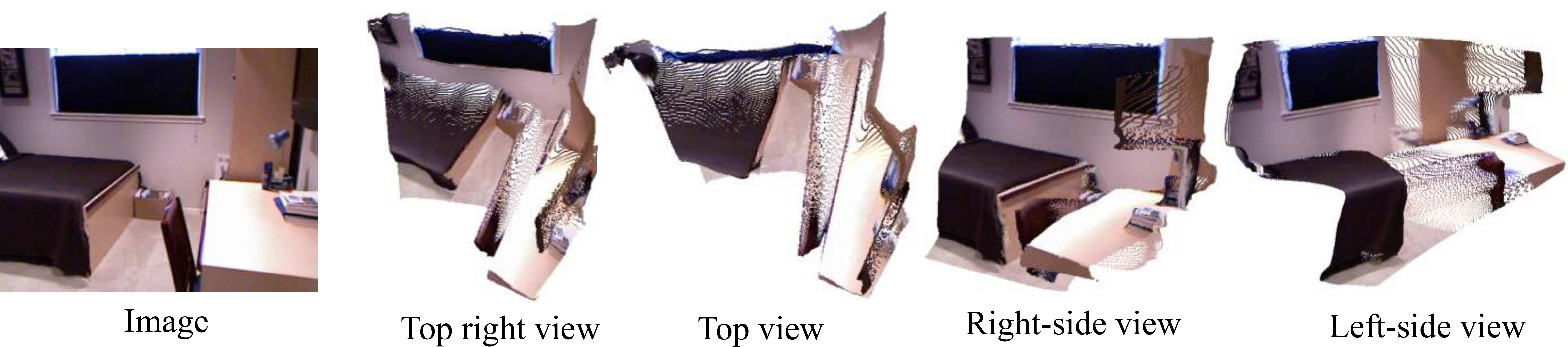}
   \vspace{2cm}
\end{minipage}

\begin{minipage}{0.99\textwidth}
\centering
   \includegraphics[width=0.99\linewidth]{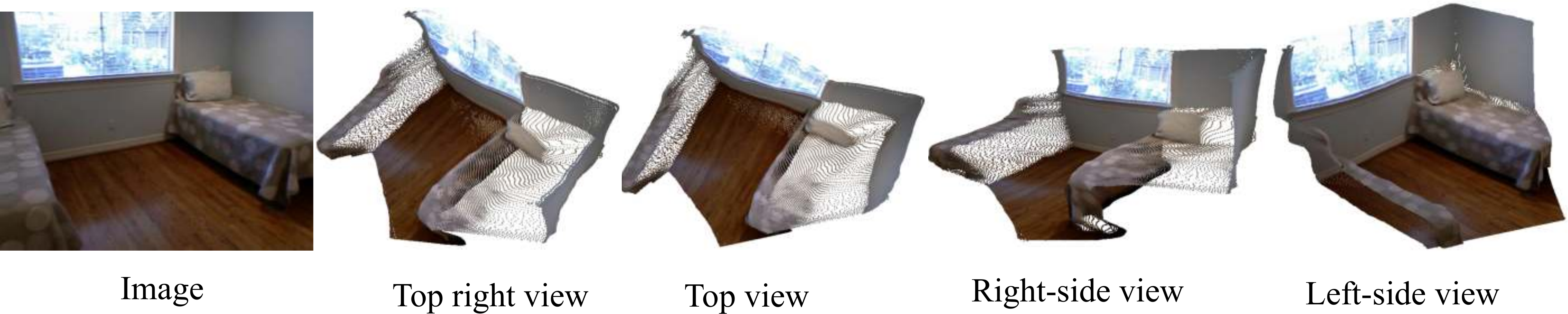}
   \vspace{2cm}
\end{minipage}
\begin{minipage}{0.99\textwidth}
\centering
   \includegraphics[width=0.99\linewidth]{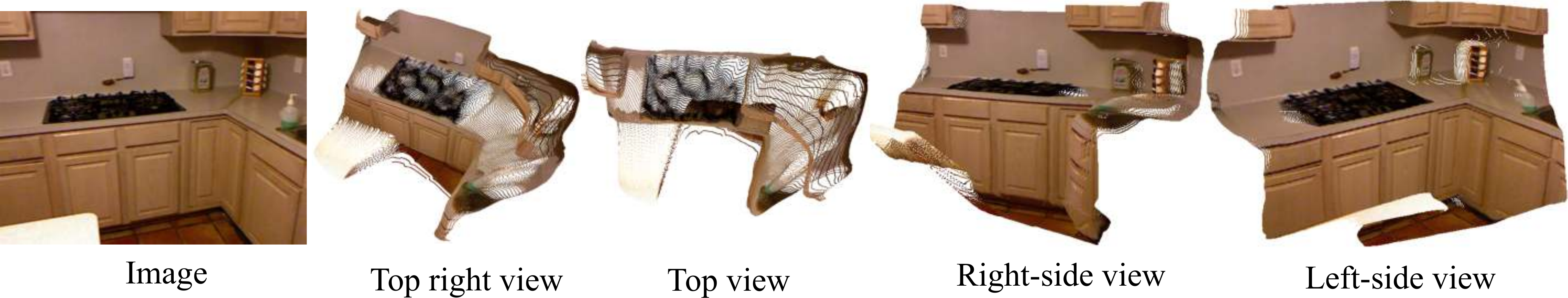}
   \vspace{1cm}
\end{minipage}
\caption{\textbf{Additional reconstruction examples from NYU Depth-v2.}}
\label{fig:nyu_3d_viz}
\end{figure*}

\end{document}